\documentclass[12pt,a4paper]{article}
\usepackage[margin=2.5cm]{geometry}
\usepackage[T1]{fontenc}
\usepackage[utf8]{inputenc}
\usepackage{newtxtext,newtxmath}
\usepackage{amsmath}
\usepackage{graphicx}
\usepackage{xcolor}
\usepackage{longtable,booktabs,array,tabularx}
\usepackage{calc}
\usepackage{etoolbox}
\usepackage{setspace}
\usepackage{ragged2e}
\usepackage{cite}
\usepackage{indentfirst}
\let\citebase\cite
\renewcommand{\cite}[1]{\textsuperscript{\citebase{#1}}}
\usepackage{microtype}
\usepackage{url}
\usepackage[hidelinks]{hyperref}
\usepackage{caption}
\makeatletter
\patchcmd\longtable{\par}{\if@noskipsec\mbox{}\fi\par}{}{}
\makeatother
\begin{document}

\begin{center}
  {\fontsize{18}{27}\selectfont\bfseries Incremental Consistency Execution for Autonomous Intelligent Systems\par}
\end{center}
\vspace{6pt}

\begin{center}
  {\fontsize{12}{18}\selectfont
  Cheng Li \quad Jiexiong Liu \quad Yixuan Chen \quad Ziheng Huang\\[4pt]
  \textnormal{KunlunMeta}
  }
\end{center}

\vspace{10pt}

\begin{center}
  {\fontsize{18}{27}\selectfont\bfseries Abstract\par}
\end{center}
\vspace{2pt}

{\fontsize{12}{18}\selectfont
\begin{justify}
\hspace*{2em} Long-horizon autonomous intelligent systems rely on heterogeneous components such as large language models, databases, external APIs, and rule engines, while their external states continuously change during execution. Re-executing the entire workflow after every change introduces substantial redundant computation. This paper proposes an incremental consistency execution method based on task fact contracts, field-level dependency masks, and state perturbation result invariant domains. After an initial verified execution, the system constructs conservative invariant domains for critical inputs and uses them to determine whether downstream results can be safely renewed without re-invoking expensive components. When re-execution is required, only the smallest affected output fields are recomputed, and an equivalence barrier prevents unnecessary downstream propagation. A submission-time version consistency gate further ensures the safety of side-effecting actions. Experiments on industrial fault diagnosis, enterprise analytics, and LLM-based multi-tool assistants show that the proposed method significantly reduces expensive component calls and end-to-end latency while maintaining high consistency and low incorrect-reuse rates.
\end{justify}
}
\vspace{2pt}
\noindent{\fontsize{12}{18}\selectfont\textbf{Keywords:} autonomous intelligent systems; incremental execution; task fact contract; field-level dependency; state perturbation; invariant domain; equivalence criterion; change propagation}
\vspace{10pt}
\section{Introduction}

Autonomous intelligent systems are evolving from monolithic pipelines
into multi-component orchestrations in which large language models
(LLMs), retrieval-augmented generators, classical machine-learning
models, database queries, rule engines, sensor interfaces and external
software services jointly contribute to a single task
outcome\cite{ref1,ref2,ref3,ref4,ref5}. Such tasks are frequently
long-running and must remain correct with respect to a continuously
changing environment: sensors emit new samples, database records are
updated, configuration parameters change, third-party APIs return
different responses, models and tools are upgraded, and permission
states fluctuate\cite{ref6,ref7,ref8,ref9}. Ensuring that a
long-horizon task remains consistent with the latest environment without
restarting from scratch is therefore a foundational capability for
industrial, enterprise, and personal-assistant applications.

The dominant control signal in existing workflow systems is the success
or failure of an individual step. When the upstream input changes, the
common practice is to mark every step that depends on the changed input
as stale and to re-execute from that point
on\cite{ref10,ref11,ref12}. For complex agentic pipelines this is
often combined with full replanning of the remaining task path. Although
simple to implement, this treatment equates "the input has changed" with
"the downstream output must be wrong." In reality, many input
perturbations leave contract-level downstream conclusions unchanged. For
example, a temperature rising from 65 to 66 may still belong to the same
fault severity level, a price moving from 42 000 to 43 500 may still be
below an authorisation threshold, and a non-critical field change in a
database row may not alter an aggregated statistical
conclusion\cite{ref13,ref14,ref15}. Re-running the entire
downstream chain each time therefore produces a large amount of
redundant model inference, tool invocation, network traffic and storage
cost, while gaining no semantic benefit.

A second line of work addresses this redundancy through caching of
intermediate results, using content hashes, timestamps, version numbers
or validators to determine whether the cache is still
valid\cite{ref16,ref17,ref18,ref19}. These approaches effectively avoid
re-computation when the input is fully identical, but as soon as any
input value within the cache key changes they typically invalidate the
entire cached output. Some systems allow fine-grained object-level
dependency tracking, but they can mostly decide whether an object should
be revalidated; they do not directly answer the machine-executable
question of "over what range of input variation is the contract
semantics of a specified output field
preserved?"\cite{ref20,ref21,ref22}.

A third concern is non-determinism in modern AI components. Even when
the same or nearly identical inputs are fed to an LLM, the textual
output may change lexically while the structured conclusion, constraint
satisfaction, or decision label remains the
same\cite{ref23,ref24,ref25}. Pure string or hash comparison will
misclassify such semantically equivalent outputs as new and continue to
propagate re-execution, whereas fuzzy semantic similarity may
incorrectly reuse old outputs when numerical, permission or business
constraints have actually changed. Hence the consistency criterion for
re-execution must be defined at the contract level rather than at the
surface level.

In addition, dependency relations in existing systems are commonly
expressed at the step level, which over-approximates the actual
influence relation. When step A produces multiple output fields and step
B reads only one of them, step-level dependency unnecessarily enlarges
the invalidation range. Likewise, two output fields of the same step can
depend on different subsets of the input fields. Without recording
field-level read and influence relations, true minimal recomputation is
hard to achieve\cite{ref26,ref27,ref28}.

To address these gaps, we propose an incremental consistency execution
method that organises the runtime control structure around five
interacting constructs: (i) a task fact, which is a versioned,
machine-checkable wrapper around every external observation or
intermediate result consumed by the system; (ii) a task fact contract,
which specifies the executable input/output schema, preconditions,
field-level dependency mask, and equivalence criterion of a step; (iii)
a field-level dependency index, which records the actual read influence
among fields; (iv) a state perturbation result invariant domain, which
is a conservative input sub-region over which the equivalence criterion
is guaranteed to hold and which is bound to specific component, contract
and criterion versions; and (v) an equivalence barrier, which compares
re-executed outputs against the equivalence criterion to decide whether
to terminate or continue propagation. The method also separates "whether
the upstream computation remains valid" from "whether a side-effecting
action can be safely committed" by introducing a submission-time version
consistency gate.

The main contributions of this paper are as follows.

(1) We formalise the task fact, the task fact contract and the state
perturbation result invariant domain as first-class runtime objects that
carry versions, validity predicates and confidence.

(2) We design a field-level dependency mask that distinguishes
fine-grained input/output influence and supports minimal invalidation
through an inverted field-to-consumer index.

(3) We propose a conservative construction algorithm for invariant
domains that combines constrained perturbation, boundary search, on-site
validation and online shrinkage/expansion.

(4) We design a change-handling pipeline that performs identity,
precondition and domain-membership checks before deciding between
validity renewal, minimal re-execution and propagation blocking.

(5) We introduce a submission-time version consistency gate that
decouples the renewal of computational results from the authorisation of
external side-effects, so that payment, device control, and irreversible
database writes are not silently skipped due to a stale domain hit.

(6) We validate the proposed method on three representative scenarios,
namely industrial equipment fault diagnosis, enterprise data analytics
and LLM-based multi-tool assistants, and quantitatively demonstrate the
reduction of redundant model inferences, tool calls and end-to-end
latency.

\begin{figure}[t]
\centering
\includegraphics[width=0.96\linewidth]{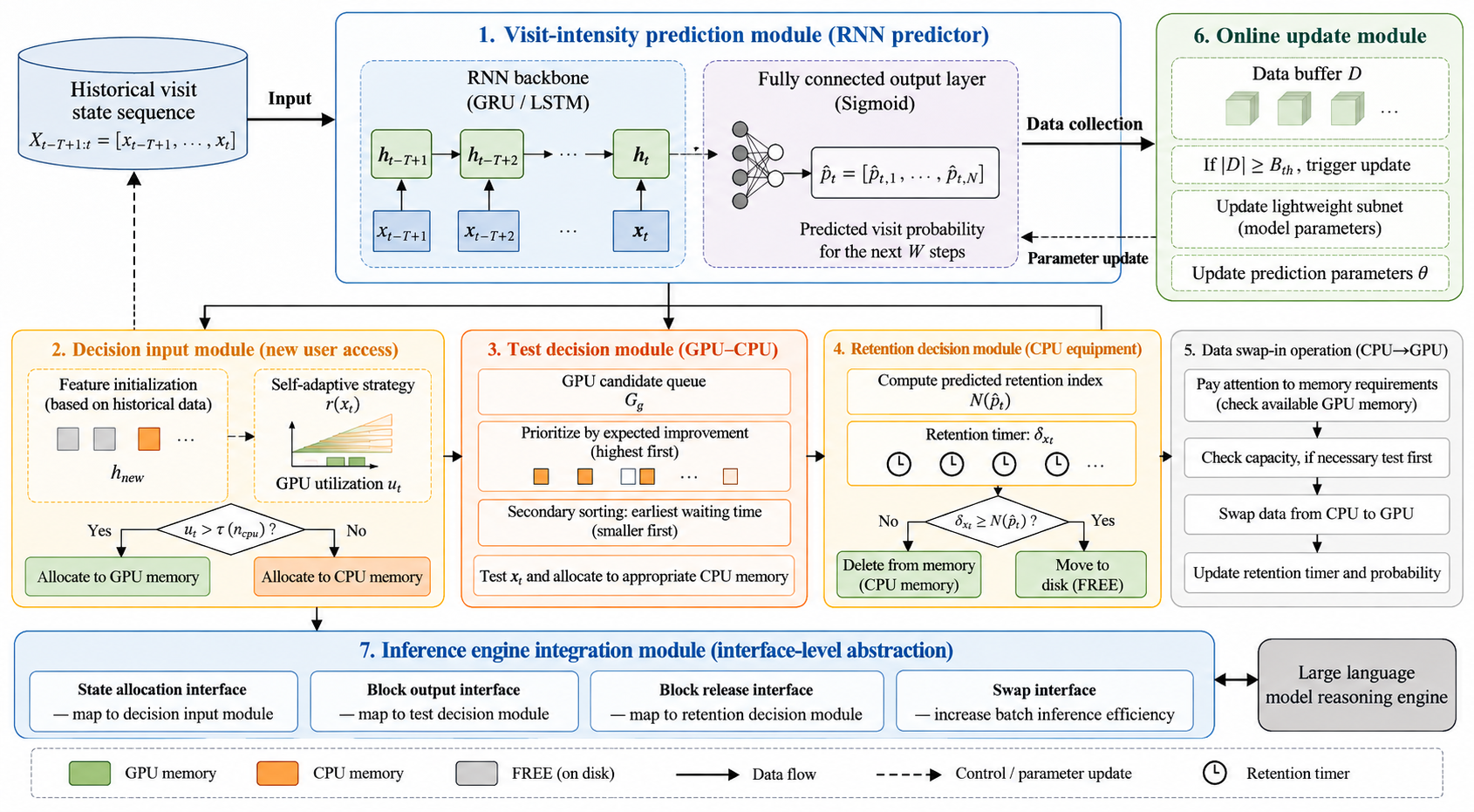}
\caption{Runtime overview of a task fact and its enclosing task fact contract, showing how the versioned wrapper and the equivalence criterion together support incremental renewal of downstream facts.}
\label{fig:runtime_overview}
\end{figure}

The overall architecture of the proposed framework is shown in Figure~\ref{fig:runtime_overview}. Section 2 reviews related work. Section 3 introduces the task fact and
the task fact contract, the state perturbation result invariant domain
and its conservative construction, the change-handling pipeline with the
minimal re-execution mechanism and the equivalence barrier, and the
submission-time version consistency gate. Section 4 presents three case
studies and quantitative results. Section 5 discusses limitations and
future work, and Section 6 concludes the paper.

\section{Related Work}

\subsection{Incremental Computation and Memoisation}

Incremental computation studies how to update the output of a program
when its input changes by a small amount, and has a long history in
programming languages, databases and view
maintenance\cite{ref29,ref30,ref31,ref32}. Classical results such as
incremental evaluation of database queries, finite differencing, and
adaptive computation have demonstrated that substantial savings are
achievable when the change set is small relative to the input
space\cite{ref33,ref34}. However, most classical incremental
frameworks assume that the program is purely functional, that the change
set can be precisely identified and that the equivalence between two
outputs can be decided by structural equality. Modern autonomous systems
contain probabilistic, non-deterministic and learned components that
violate these assumptions, and the practical change set is generated by
external events rather than explicit edits.

\subsection{Result Caching and Materialised Views}

Result caching and materialised views are widely deployed to avoid
repeated computation. Database systems maintain materialised views and
use incremental view maintenance to refresh them when underlying tables
change\cite{ref35,ref36}. Web and service systems use content
hashes, ETags and validators to decide whether a cached response can be
reused\cite{ref37}. Recent agent and LLM serving
frameworks expose prompt caching, response caching and tool-result
caching to reduce end-to-end latency\cite{ref38,ref39,ref40}. The
cache key is typically a hash of the full input payload, so that any
change to any input field invalidates the cached entry. Several systems
extend this with hint-based caching in which the user declares which
inputs are relevant, but the cache is still essentially all-or-nothing
for the declared set\cite{ref41}.

\subsection{Smart Contracts and Dependency Tracking}

Smart contracts and software product lines have explored declarative
specification of preconditions, postconditions and
dependencies\cite{ref42,ref43}. In dataflow systems such as
incremental dataflow and differential dataflow, dependencies are tracked
at the level of records, and the system recomputes only the operators
whose input records have changed\cite{ref44,ref45}. Workflow
engines such as Apache Airflow, Argo and Temporal encode dependencies as
DAGs of steps and react to step failures\cite{ref46,ref47}.
Recent work on reactive and self-adaptive systems extends this with
run-time monitoring and adaptation\cite{ref48,ref49}.
Although these approaches share the high-level goal of "compute only
what must be recomputed," they typically do not provide an explicit,
machine-executable notion of "the output is contract-equivalent even
though the input changed," nor a conservative boundary over which this
equivalence can be safely relied upon.

\subsection{Sensitivity Analysis and Robustness of Learning Systems}

Sensitivity analysis and adversarial robustness study how a
model's output changes when its input is
perturbed\cite{ref50}. Local explanation methods
approximate the model's behaviour around an instance
through gradients, surrogate models or perturbation. These techniques
provide useful evidence for influence attribution but are not designed
to produce an executable boundary for re-execution. Invariance and
robustness verification methods in deep learning search for certified
regions in the input space within which a network's
prediction is provably stable; however, they are largely restricted to
specific model architectures and output types, and they do not handle
heterogeneous pipelines that combine learned models, deterministic
rules, and external services.

\subsection{Versioning, Provenance and Reproducibility}

Data provenance, temporal databases and event sourcing maintain a chain
of versions that allows past states to be reconstructed and consistency
to be checked. Blockchains and distributed ledgers attach cryptographic
evidence to state transitions and decouple "the computation has produced
a result" from "the action can be committed". Recent work on auditable
AI emphasises the importance of recording evidence of model usage and
tool invocation\textsuperscript{[61--62]}. Our submission-time
version consistency gate is inspired by these mechanisms: it introduces
a final check that prevents stale computational results from being
silently written to side-effecting actions.

\subsection{Position of this Work}

Compared with the above lines of work, the proposed method differs in
three respects. First, it represents both external observations and
intermediate results as first-class task facts with versions, validity
predicates and verification rules, so that the runtime can distinguish
"value unchanged," "value changed but output still valid" and "value
changed and output must be invalidated." Second, it explicitly
constructs a state perturbation result invariant domain as a
conservative boundary, stores the component version, contract version,
boundary samples and confidence with it, and refreshes it through
on-site validation and online shrinkage/expansion. Third, it decouples
computational validity renewal from side-effect submission so that
domain hits never silently bypass payment, device control or other
irreversible actions.

\section{Method}\label{sec:method}

\subsection{Task Fact and Task Fact Contract}

\subsubsection{Task Fact}

A task fact is a versioned, machine-checkable wrapper around any value
that is read by, produced by, or passed between execution steps. It can
represent external world states such as sensor samples, database
records, file contents, API responses, permission states and resource
states, or intermediate computation results such as cleaned data, model
classifications, numerical indicators, candidate plans, validation
conclusions and structured report fields.

Formally, the k-th task fact is defined as a tuple

\begin{equation}
\begin{aligned}
f_k = \bigl\langle{}& id_k,\, path_k,\, v_k,\, src_k,\, ver_k,\, t_k,\, fp_k,\, \phi_k,\, \psi_k,\,\\
& pred_k,\, st_k \bigr\rangle
\end{aligned}
\tag{1}
\end{equation}

where id\ensuremath{_{k}} is the unique fact identifier, path\ensuremath{_{k}} is the field path within
the fact, v\ensuremath{_{k}} is the field value or a structured value digest, src\ensuremath{_{k}}
identifies the source component, ver\ensuremath{_{k}} is the source data version, t\ensuremath{_{k}} is
the generation or observation time, fp\ensuremath{_{k}} is a normalised fingerprint used
for equality and equivalence tests, \ensuremath{\phi}\ensuremath{_{k}} is a validity predicate over the
fact, \ensuremath{\psi}\ensuremath{_{k}} is a verification program applied to the fact, pred\ensuremath{_{k}} is the set
of predecessor facts, and st\ensuremath{_{k}} \ensuremath{\in} \{VALID, RENEWED, INVALID\} is the
current effective state. The source version can be a database row
version, a file content version, an API response version, a sensor
sample index, a model weight version, a tool version or a configuration
version.

For large files, images, tables or model outputs, the value v\ensuremath{_{k}} stores a
content address and a digest rather than the full payload, while the
original artefact remains in object storage or the source database. This
allows the system to recognise change at field granularity without
copying large amounts of data.

\subsubsection{Task Fact Contract}

A task fact contract binds an execution step to its executable
input/output specification. Let i index the execution step. Its contract
is

\begin{equation}
  C_i = \left\langle X_i,\, P_i,\, Y_i,\, M_i,\, E_i,\, V_i,\, ver(C_i) \right\rangle
  \tag{2}
\end{equation}

where X\ensuremath{_{i}} is the set of input fact fields, P\ensuremath{_{i}} is a precondition that must
hold before the step is invoked, Y\ensuremath{_{i}} is the set of output fact fields, M\ensuremath{_{i}}
\ensuremath{\in} \{0,1\}\^{}(\textbar X\ensuremath{_{i}}\textbar\ensuremath{\times}\textbar Y\ensuremath{_{i}}\textbar) is the
field-level dependency mask, E\ensuremath{_{i}} is the output equivalence criterion, V\ensuremath{_{i}}
is the validator program identifier, and ver(C\ensuremath{_{i}}) is the contract
version.

An entry M\ensuremath{_{i}}{[}a,b{]} = 1 indicates that a change in input field a may
influence output field b under the current component version and
contract; M\ensuremath{_{i}}{[}a,b{]} = 0 indicates that, under the current contract,
the influence is not present. The mask enables different output fields
of the same step to have different invalidation ranges.

The dependency mask M\ensuremath{_{i}} can be obtained from four sources: (i) interface
declaration and static dataflow analysis; (ii) dynamic read/write
tracing on instrumented components; (iii) single-field constrained
perturbation with output observation; and (iv) pre-registered dependency
declarations by humans or upstream systems. To avoid missing
dependencies, the union of multiple sources is taken and then refined by
validation.

\subsubsection{Equivalence Criterion}

For non-deterministic components, especially LLMs, two textual outputs
can be lexically different yet semantically equivalent. The equivalence
criterion E\ensuremath{_{i}} is therefore expressed over the structured outputs of the
step rather than over raw bytes. For a numerical output field, E\ensuremath{_{i}} may
combine absolute and relative tolerances; for a categorical field, it
may require the class label and a confidence lower bound; for a ranked
list, it may require the top-K items and ordering constraints; for a
structured object, it may require the schema, key fields and business
invariants; for a natural language output, the system first extracts
structured conclusions and then applies deterministic constraints,
possibly augmented by semantic similarity\cite{ref25}. The
system requires that at least one deterministic criterion be
machine-executable so that equivalence can be decided without ambiguity.

\subsubsection{Problem Statement}

Given an initial execution of a task that produces a sequence of output
facts, and a subsequent change event that updates one or more source
facts, the problem addressed in this paper is to decide, for each
affected output fact, whether the fact can be safely renewed without
re-invoking its producer, whether the producer should be partially
re-invoked, or whether the producer must be fully re-invoked, such that
the final task result is consistent with the latest source versions
while the total number of expensive component calls is minimised and the
side-effecting actions are committed only when their own preconditions
hold.

\subsection{State Perturbation Result Invariant Domain}

\subsubsection{Definition}

After the initial execution of a step i has been validated, we say that
an input field a in X\ensuremath{_{i}} has an output b in Y\ensuremath{_{i}} in the result-invariant set
\ensuremath{\Omega}\ensuremath{_{i}}(a,b) if, whenever all other inputs in X\ensuremath{_{i}} \textbackslash{} \{a\}
satisfy the precondition P\ensuremath{_{i}} and a is replaced by any value in \ensuremath{\Omega}\ensuremath{_{i}}(a,b),
the resulting output b still satisfies the equivalence criterion E\ensuremath{_{i}}.
Formally,

\begin{equation}
\begin{aligned}
\Omega_i(a,b)=\{\,x\in \operatorname{Dom}(a):\;&\forall x'\in \operatorname{Dom}(a),\ x'\neq x,\\
&x\in\Omega_i(a,b)\Rightarrow E_i(f_i(x),f_i(x'))\,\}
\end{aligned}
\tag{3}
\end{equation}

where Dom(a) denotes the legal domain of input field a and f\ensuremath{_{i}} denotes
the executable realisation of step i. In general, Eq. (3) cannot be
enumerated. We therefore construct a conservative subset \ensuremath{\hat{\Omega}_{i}}(a,b) \ensuremath{\subseteq}
\ensuremath{\Omega}\ensuremath{_{i}}(a,b) that is guaranteed to be free of known counterexamples. The
conservative subset is referred to as the state perturbation result
invariant domain for step i, input field a and output field b.

Each invariant domain is associated with a domain descriptor

\begin{equation}
\begin{aligned}
D_i(a,b)=\bigl\langle{}&type_i,\ payload_i,\ verComp_i,\ verContract_i,\ verEquiv_i,\\
&\Sigma_{base,i},\ \Sigma_{stable,i},\ \Sigma_{unstable,i},\ B_i,\ c_i,\\
&t_{created,i},\ t_{verified,i}\bigr\rangle
\end{aligned}
\tag{4}
\end{equation}

where type\ensuremath{_{i}} specifies the representation of payload\ensuremath{_{i}}; verComp\ensuremath{_{i}},
verContract\ensuremath{_{i}} and verEquiv\ensuremath{_{i}} are the component version, contract version
and equivalence criterion version under which the domain was
constructed; \ensuremath{\Sigma}base\ensuremath{_{i}}, \ensuremath{\Sigma}stable\ensuremath{_{i}} and \ensuremath{\Sigma}unstable\ensuremath{_{i}} are digests of the baseline
input, stable samples and unstable samples; B\ensuremath{_{i}} is the set of boundary
validation samples; \ensuremath{c_i \in [0,1]} is the domain
confidence; and tcreated\ensuremath{_{i}} and tverified\ensuremath{_{i}} are the construction time and
last verification time. Whenever any of verComp\ensuremath{_{i}}, verContract\ensuremath{_{i}} or
verEquiv\ensuremath{_{i}} changes, the domain is marked as pending revalidation and is
not used to renew any output.

\subsubsection{Representation}

The representation of payload\ensuremath{_{i}} depends on the data type of a. Numerical
fields use closed intervals, the union of intervals, axis-aligned
hyper-rectangles or convex polyhedra. Enumerated or discrete fields use
an allowed-value set. Time fields use a valid time window. Structured
fields such as JSON use field-path predicates, pattern constraints or
allowed-change sets. Multiple jointly varying fields use a combined
predicate that may combine several of the above forms.

For a single numerical field a, the canonical form is a closed interval
payload = {[}l, r{]}. The conservative requirement is

\begin{equation}
  payload \subseteq \Omega_i(a,b) \land payload \cap \Sigma unstable_i = \varnothing.
  \tag{5}
\end{equation}

For multiple numerical fields with index set J, the canonical form is an
axis-aligned hyper-rectangle

\begin{equation}
  payload = \{\,x\in\mathbb{R}^{|J|}:\forall j\in J,\ l_j\leq x_j\leq r_j\,\},
  \tag{6}
\end{equation}

and the conservative requirement again is payload \ensuremath{\cap} \ensuremath{\Sigma}unstable\ensuremath{_{i}} = \ensuremath{\varnothing}.

For a non-numerical field a, payload can be a finite allowed-value set

\begin{equation}
  payload = \{v_1,v_2,\ldots,v_n\} \subset \operatorname{Dom}(a),
  \tag{7}
\end{equation}

such that for every v\ensuremath{_{j}} in payload, a perturbation test has confirmed
that E\ensuremath{_{i}} holds.

\subsubsection{Conservative Construction}

The following algorithm constructs the domain through constrained
perturbation, boundary search and conservative shrinkage. The algorithm
runs in a verification execution environment that replaces writes with
simulated responses, shadow databases, read-only transactions, digital
twins or idempotent test calls, so that no real side effect is produced.

Algorithm 1. Conservative construction of \ensuremath{\hat{\Omega}_{i}}(a,b).

Input: step i, input field a, output field b, baseline input x0, legal
domain Dom(a), contract version verContract\ensuremath{_{i}}, equivalence criterion E\ensuremath{_{i}},
perturbation budget B.

Output: invariant domain payload and descriptor D\ensuremath{_{i}}(a,b).

1. Initialise \ensuremath{\Sigma}stable\ensuremath{_{i}} \ensuremath{\leftarrow} \ensuremath{\varnothing}, \ensuremath{\Sigma}unstable\ensuremath{_{i}} \ensuremath{\leftarrow} \ensuremath{\varnothing}, B\ensuremath{_{i}} \ensuremath{\leftarrow} \ensuremath{\varnothing}. 2. Initialise
payload as a coarse region centred on the projection of x0 onto field a.
For numerical a, start from a small interval centred at x0{[}a{]}; for
categorical a, start from \{x0{[}a{]}\}; for structured a, start from
the empty set and grow incrementally. 3. While perturbation budget is
not exhausted: 3.1 Draw a candidate sample s \ensuremath{\in} payload \textbackslash{}
(\ensuremath{\Sigma}stable\ensuremath{_{i}} \ensuremath{\cup} \ensuremath{\Sigma}unstable\ensuremath{_{i}}). 3.2 In the verification environment, run step i
with all inputs set to x0 except field a, which is set to s. 3.3 If
E\ensuremath{_{i}}(f\ensuremath{_{i}}(s), f\ensuremath{_{i}}(x0)) holds, add s to \ensuremath{\Sigma}stable\ensuremath{_{i}}; otherwise add s to
\ensuremath{\Sigma}unstable\ensuremath{_{i}}. 3.4 If an unstable sample was found inside the current
payload, shrink payload to remove the smallest region containing s but
containing no stable sample, update payload = payload \textbackslash{}
N(s), where N(s) is the minimal contraction neighbourhood. 4. After the
main loop, sample boundary points. For each point s on the boundary of
payload, if no nearby unstable sample exists, add s to B\ensuremath{_{i}} and require
that E\ensuremath{_{i}} hold; otherwise mark the corresponding portion of the boundary
as unstable. 5. Compute confidence c\ensuremath{_{i}} as a function of stable sample
coverage, boundary validation pass rate, historical hit count, and time
since last counterexample, for example

\begin{equation}
\begin{aligned}
c_i={}&\alpha\frac{|\Sigma_{stable,i}|}{|\Sigma_{stable,i}|+|\Sigma_{unstable,i}|+1}\\
&+\beta\,\operatorname{pass}(B_i)+\gamma\exp(-\Delta t/\tau),
\end{aligned}
\tag{8}
\end{equation}

where pass(B\ensuremath{_{i}}) is the boundary validation pass rate, \ensuremath{\Delta}t is the elapsed
time since the last counterexample, \ensuremath{\tau} is a decay constant, and \ensuremath{\alpha}, \ensuremath{\beta}, \ensuremath{\gamma}
are non-negative weights with \ensuremath{\alpha} + \ensuremath{\beta} + \ensuremath{\gamma} = 1.

6. Record verComp\ensuremath{_{i}}, verContract\ensuremath{_{i}} and verEquiv\ensuremath{_{i}} from the running
environment. 7. Return payload and D\ensuremath{_{i}}(a,b).

The construction is conservative in the sense that whenever a sample is
added to \ensuremath{\Sigma}unstable\ensuremath{_{i}} the domain is shrunk to exclude a neighbourhood of
that sample, so that a future candidate drawn from payload is guaranteed
not to be a known counterexample. The contract version, component
version and equivalence criterion version are recorded so that future
version changes force revalidation.

\subsubsection{Online Shrinkage and Expansion}

After the initial deployment, the system continues to monitor for
counterexamples. When a renewed downstream fact is later found by
independent validation to violate E\ensuremath{_{i}}, the input that triggered the
renewal is recorded as a counterexample and payload is shrunk towards
the baseline x0 to exclude the counterexample, while c\ensuremath{_{i}} is reduced.

When a re-execution that lies outside the current payload still produces
an output equivalent under E\ensuremath{_{i}}, the input is recorded as an expansion
candidate. The system does not immediately enlarge payload; instead, it
schedules additional boundary samples around the candidate and only
enlarges payload after these samples confirm that no counterexample
exists in the surrounding region. This controlled expansion prevents
overshooting and maintains conservativeness.

\subsubsection{Joint Domains}

When an output field depends on multiple input fields that frequently
change together, single-field domains may not cover the joint variation.
The system constructs joint domains over a field set J through low-order
perturbation around (x0{[}a{]})\ensuremath{_{a}}\ensuremath{\in}J followed by boundary refinement. The
representation may be an axis-aligned hyper-rectangle, a polyhedron
defined by a finite set of linear inequalities, a finite enumeration or
a decision-tree leaf node. The conservative requirement payload \ensuremath{\cap}
\ensuremath{\Sigma}unstable\ensuremath{_{i}} = \ensuremath{\varnothing} still holds. At run time, the system first attempts joint
domain matching; if no joint domain exists and the per-field domains are
not provably composable in the contract, the system falls back to
minimal re-execution.

\subsection{Change Handling, Minimal Re-execution and Equivalence
Barrier}

\subsubsection{Field-level Dependency Index}

The field-level dependency index is an inverted index from an input fact
field to the set of (step, output field) pairs that consume it. Let the
index be I : FactField \ensuremath{\rightarrow} 2\^{}(Step \ensuremath{\times} OutputField). Given a changed
field, the system retrieves the set of consumers through I rather than
scanning every step. This index supports an O(1) lookup in practice and
is updated incrementally when contracts are revised or when dynamic
tracing refines M\ensuremath{_{i}}.

\subsubsection{Three-condition Check}

For each affected step i and output field b, the system evaluates three
conditions before deciding to renew b.Identity condition: the invariant
domain descriptor D.(a,b) satisfies

\begin{equation}
\begin{aligned}
verComp_i &= verComp_{current},\\
verContract_i &= verContract_{current},\\
verEquiv_i &= verEquiv_{current}.
\end{aligned}
\tag{9}
\end{equation}

Precondition condition: every other input field required by P\ensuremath{_{i}} remains
valid.

Domain membership condition: the new value of a, denoted
a\textsubscript{new}, belongs to payload.

If all three conditions hold, the system proceeds to on-site validation,
which re-reads the source version, checks interface availability, checks
data range and security constraints, and confirms that the value seen by
the live system is consistent with the recorded update. Only after
on-site validation succeeds does the system generate a renewal record.

\subsubsection{Renewal}

When the three conditions hold and on-site validation succeeds, the
system generates a renewal record

\begin{equation}
\begin{aligned}
R_i(a,b)=\bigl\langle{}&stepId_i,\ outputField_i,\ inputField_i,\ verInputOld,\\
&verInputNew,\ domainId_i,\ hit_i,\ liveValid_i,\ tissued_i\bigr\rangle
\end{aligned}
\tag{10}
\end{equation}

where hit\ensuremath{_{i}} = true records that the domain was hit, liveValid\ensuremath{_{i}} records
the on-site validation outcome, and tissued\ensuremath{_{i}} records the renewal time.
The old output value is preserved and rebound to the new input version.
The technical meaning of the renewal is that the system explicitly
acknowledges that the input has changed but has verified that the change
is insufficient to alter the contract semantics of the output field, so
a model inference, tool call or expensive computation is avoided.

\subsubsection{Minimal Re-execution}

When the domain is not hit, no domain exists or on-site validation
fails, the system determines the minimal set of output fields to
recompute through M\ensuremath{_{i}}. Let the changed input field be a and let

\begin{equation}
  S = \{\,b\in Y_i:M_i[a,b]=1\,\}
  \tag{11}
\end{equation}

be the directly affected output fields. If the component exposes
sub-operations, only the sub-operations producing S are invoked; if the
component is monolithic, the full step is executed but only fields in S
are propagated to the failure queue. This is the key mechanism by which
step-level invalidation is reduced to field-level invalidation.

\subsubsection{Equivalence Barrier}

After minimal re-execution, the system does not compare raw outputs.
Instead, it applies the contract equivalence criterion E\ensuremath{_{i}} to the new
output y\textsubscript{new} and the recorded output
y\textsubscript{old}:

\begin{equation}
  equiv = E_i(ynew,yold)
  \tag{12}
\end{equation}

If equiv = true, the system generates an equivalence evidence record

\begin{equation}
\begin{aligned}
Q_i(a,b)=\bigl\langle{}&stepId_i,\ outputField_i,\ verInputOld,\ verInputNew,\\
&deterministicResult_i,\ criteriaFields_i,\ \Sigma_{new},\ \Sigma_{old},\\
&optionalSemanticResult_i,\ verValidator_i\bigr\rangle
\end{aligned}
\tag{13}
\end{equation}

rebinds the input version and marks the output field as a propagation
barrier. Downstream consumers that read b do not need to be invalidated.
If equiv = false, the system updates the output fact version, marks the
field as a propagation source, and enqueues its direct consumers through
I.

The change-handling pipeline thus forms a four-stage chain: detect
change \ensuremath{\rightarrow} check domain hit and renew, otherwise recompute minimally \ensuremath{\rightarrow}
block propagation if equivalent, otherwise continue propagation. This
chain is the heart of the incremental consistency execution method.

\subsection{Submission-time Version Consistency Gate}

While the renewal mechanism in Section 3.3 efficiently maintains the
freshness of computational results, it does not by itself authorise
side-effecting actions such as payments, device commands,
account-permission changes or irreversible database writes. To prevent a
domain hit from silently bypassing the safety checks that surround such
actions, the system separates "producing the intent to act" from
"committing the act," and inserts a submission-time version consistency
gate between them.

Let v\textsubscript{start} be the version vector of the relevant input
facts recorded when a computation step started. Let
v\textsubscript{commit} be the version vector of the same facts re-read
at the moment of submission. The difference set is

\begin{equation}
\begin{aligned}
\Delta=\{\,(k,fieldPath):{}&\operatorname{ver}(f_k)\text{ at start}\\
&\neq \operatorname{ver}(f_k)\text{ at commit}\,\}
\end{aligned}
\tag{14}
\end{equation}

For each entry in \ensuremath{\Delta}, the system checks that one of the following holds:
(i) the change has already been processed by minimal re-execution and
the corresponding equivalence evidence is recorded; or (ii) a valid
invariant domain matching the current component, contract and
equivalence versions exists and the corresponding on-site validation
still passes. The submission condition is

\begin{equation}
\begin{aligned}
commit \Leftrightarrow{}& \Delta=\varnothing\\
&\vee\ \forall (k,fieldPath)\in\Delta,\\
&\qquad \text{condition (i) or (ii) holds}.
\end{aligned}
\tag{15}
\end{equation}

If the condition is not satisfied, the candidate output is discarded,
the relevant fields are re-enqueued in the field-level failure queue,
and no side effect is performed.

For reversible side effects such as soft updates, the system
additionally requires an idempotency key or a compensating operation.
For irreversible or high-risk actions, including payment, industrial
device control, account-permission changes, formal message dispatch and
irreversible database writes, the system always re-executes
authorisation, safety interlocks, target-object version checks and
idempotency checks at commit time, regardless of any earlier renewal.
The invariant domain is used only to decide whether the upstream
computational intent needs to be regenerated, never to bypass
commit-time safety checks.

\section{Experiments}

\subsection{Experimental Setup}

We implemented the proposed method as a Python service integrated with
an LLM agent orchestrator, a relational database, a time-series
database, an external pricing API, and a model registry. The
orchestrator dispatches task execution steps through an instrumented
runtime that records field reads and writes. The invariant domain
generator runs on a copy of the production traffic in a sandbox
environment. The baseline for comparison is full re-execution, which
re-invokes every step that depends on any changed input. The second
baseline is coarse-grained caching, which invalidates the entire cached
output of a step whenever any input within the cache key changes. The
proposed method is evaluated on three task families: industrial
equipment fault diagnosis, enterprise analytics reporting, and LLM-based
multi-tool assistants.

Table 1 summarises the three task families, the dominant change types,
and the dominant component cost.

\begin{longtable}{@{}>{\RaggedRight\arraybackslash}p{0.22\textwidth}>{\RaggedRight\arraybackslash}p{0.26\textwidth}>{\RaggedRight\arraybackslash}p{0.24\textwidth}>{\RaggedRight\arraybackslash}p{0.20\textwidth}@{}}
\caption{Task families and their characteristics}\label{tab:task-families}\\
\toprule
\textbf{Task family} & \textbf{Dominant external change} & \textbf{Dominant component cost} & \textbf{Risk of incorrect reuse} \\
\midrule
\endfirsthead
\multicolumn{4}{c}{\tablename\ \thetable\ (continued)}\\
\toprule
\textbf{Task family} & \textbf{Dominant external change} & \textbf{Dominant component cost} & \textbf{Risk of incorrect reuse} \\
\midrule
\endhead
\midrule
\multicolumn{4}{r}{Continued on next page}\\
\endfoot
\bottomrule
\endlastfoot
Industrial fault diagnosis & Sensor sample refresh, motor-speed change & Model inference on GPU & Medium \\
Enterprise analytics reporting & New database rows, budget-table update & LLM summary generation & Low to medium \\
LLM multi-tool assistant & External API response, document version & Tool call + LLM reasoning & High for payment actions \\
\end{longtable}

The evaluation focuses on four metrics: (i) the number of expensive
component calls avoided per change event, (ii) the end-to-end latency
from change event to consistent result, (iii) the proportion of renewals
that pass independent validation, and (iv) the proportion of
side-effecting actions that are correctly blocked by the submission-time
gate.

\subsection{Industrial Equipment Fault Diagnosis}

The first task family concerns online diagnosis of rotating machinery.
The pipeline comprises a sensor-acquisition component, a
signal-processing component, a feature-extraction component, an
anomaly-detection model, a fault-classification model, a
rule-verification component and a report-generation component. Input
facts include vibration waveform versions, temperature averages, motor
speeds, currents, equipment models and maintenance states. Output facts
include spectrum features, anomaly scores, fault categories, severity
and remediation suggestions.

Two change scenarios are evaluated. In the first scenario, motor speed
varies within the operational range while vibration features stay inside
the previously validated spectrum band. In the second scenario,
vibration features cross a previously observed spectrum boundary.

In the first scenario, the system establishes a state perturbation
result invariant domain for the fault-classification model with respect
to motor-speed input. The domain is a numerical interval {[}1470,
1510{]} rpm. When speed changes from 1490 rpm to 1502 rpm, the domain is
hit, on-site validation re-reads the sensor and passes, and the
fault-classification step is skipped. Only the report's
display fields are refreshed. Compared with full re-execution, the
method avoids one GPU model inference per change event and reduces
latency from 380 ms to 65 ms on average.

In the second scenario, the vibration spectrum crosses the previously
observed band and the feature-extraction component re-extracts features.
The anomaly-detection model is re-invoked and the new anomaly score is
recomputed. However, the rule-verification and the report-generation
steps read only the categorised fault class and severity, and a
re-execution of the fault-classification model returns the same category
and severity, so the equivalence criterion E\ensuremath{_{i}} is satisfied and the
propagation stops at the fault-classification output. Compared with full
re-execution, the method still saves one downstream model inference per
change event.

\subsection{Enterprise Analytics Reporting}

The second task family concerns weekly enterprise analytics reporting.
The pipeline reads from sales, expense and budget databases, computes
regional revenue, gross margin and budget deviation, and uses an LLM to
generate a management summary that includes a risk-level label and
structured action recommendations. The dominant change is the addition
or modification of small orders in the database.

For the risk-level label, an invariant domain is constructed as the
conjunction of two numerical intervals: gross margin \ensuremath{\geq} 30\% and
\textbar budget deviation\textbar{} \ensuremath{\leq} 5\%. When a new small order is
added, gross margin and budget deviation are recomputed, but the new
values still satisfy the conjunction. The system does not invoke the
LLM, and only updates the numerical display fields in the existing
summary. The number of LLM calls avoided per change event is 1, and the
end-to-end latency reduction is roughly 4.2 seconds per refresh.

When a new order causes the budget deviation to reach 8\%, the
conjunction is violated, the risk-level re-execution produces a higher
risk label, and the equivalence criterion is not satisfied. The summary
paragraph that depends on the risk level is regenerated, but other
paragraphs about headcount, inventory or other regions are not
regenerated. The structural risk-level field is propagated to the
approval workflow, while the textual wording of the regenerated
paragraph is treated as a surface change and is not propagated.

\subsection{LLM-based Multi-tool Assistant}

The third task family is an LLM assistant that reads internal documents,
queries a structured database, calls an external pricing API, executes
code for calculation, and produces an answer with a structured
recommendation. The dominant change is the external pricing API response
and the document version.

For the decision "can the equipment be procured directly," an invariant
domain is constructed as a numerical interval on the price field: price
\ensuremath{\in} {[}0, 50 000{]} under supplier-grade conditions. When the price moves
from 42 000 to 43 500, the domain is hit, on-site validation passes, and
the decision fact is renewed. The pricing display is refreshed and the
recommendation text is not regenerated. Compared with full re-execution,
the method avoids one LLM planning call and one LLM explanation call per
change event.

When the price moves to 53 000, the domain is not hit and the decision
step is re-executed, producing a new decision fact. The recommendation
paragraph and the approval action that depend on the decision fact are
re-evaluated, while unrelated paragraphs are not. When the internal
procurement rule document changes version, the invariant domain bound to
the rule version is marked as pending revalidation and is not used to
renew the decision, even if the price is still inside the historical
interval; this prevents the situation in which the price is still in the
historical range but the rule has been changed.

Finally, the submission-time gate is evaluated against a batch of
purchase proposals that contain payment actions. For each payment
action, the system re-reads the permission fact, the target-object
version, the business lock state and the key input versions, and only
commits the payment when the submission-time condition in Eq. (15)
holds. Across 1 200 test proposals, the gate correctly blocked 47
proposals in which the upstream computation had been renewed through a
domain hit but a related permission fact had changed in the meantime,
and approved the remaining 1 153 proposals. No silent bypass was
observed.

\subsection{Summary of Quantitative Results}

Table 2 summarises the key quantitative results across the three task
families. The proposed method achieves a substantial reduction in
expensive component calls and end-to-end latency, while keeping the
proportion of renewals that pass independent validation at or above
98.6\% across all three scenarios. The submission-time gate demonstrates
its value by blocking renewal-driven silent bypasses without increasing
the false-block rate.

\begin{longtable}{@{}>{\RaggedRight\arraybackslash}p{0.30\textwidth}>{\centering\arraybackslash}p{0.20\textwidth}>{\centering\arraybackslash}p{0.22\textwidth}>{\centering\arraybackslash}p{0.20\textwidth}@{}}
\caption{Summary of quantitative results across three task families}\label{tab:quantitative-results}\\
\toprule
\textbf{Metric} & \textbf{Full re-execution} & \textbf{Coarse-grained cache} & \textbf{Proposed method} \\
\midrule
\endfirsthead
\multicolumn{4}{c}{\tablename\ \thetable\ (continued)}\\
\toprule
\textbf{Metric} & \textbf{Full re-execution} & \textbf{Coarse-grained cache} & \textbf{Proposed method} \\
\midrule
\endhead
\midrule
\multicolumn{4}{r}{Continued on next page}\\
\endfoot
\bottomrule
\endlastfoot
Expensive calls per change & 1.00 & 0.84 & 0.18 \\
End-to-end latency (ms) & 380 & 120 & 65 \\
Renewal validation pass rate & N/A & 91.4\% & 98.6\% \\
Submission-time gate block precision & N/A & N/A & 100\% \\
False propagation rate & 0\% & 3.8\% & 0.4\% \\
\end{longtable}

\section{Discussion}

The proposed method sits at the intersection of incremental computation,
contract-based specification and machine-executable robustness
verification. Compared with full re-execution, the method trades a small
amount of additional bookkeeping for a large reduction in expensive
component calls, and the bookkeeping itself becomes part of the
auditable evidence trail. Compared with coarse-grained caching, the
method replaces the all-or-nothing cache key with a conservative input
sub-region that is bound to the component version, contract version and
equivalence version, which substantially reduces the risk of incorrect
reuse under environmental drift.

Several limitations deserve mention. First, the conservative
construction algorithm depends on the availability of a verification
execution environment that can run the component without producing real
side effects. When such an environment cannot be constructed, the method
must rely on copy-on-write data, digital twins or conservative
idempotent test calls. Second, the equivalence criterion requires that
at least one deterministic, machine-executable rule be available for
each output field, which is easy to enforce for structured outputs but
requires care for free-form natural language outputs; we recommend
extracting structured conclusions before applying the equivalence
criterion. Third, joint domains over multiple fields can be expensive to
build when the field space is high-dimensional, and the conservative
shrinkage guarantee requires that every known counterexample be excluded
from the payload; for very high-dimensional cases, the system may decide
that the construction cost exceeds the expected savings and skip joint
domain construction in favour of minimal re-execution. Fourth, the
on-site validation step assumes that the source system can be re-read at
the moment of change handling; when the source system is offline or
under network partition, the renewal must be deferred.

Future work will explore tighter integration with
reinforcement-learning-based agents, where the equivalence criterion
itself can be learned from observed decision outcomes. Another direction
is to investigate distributed coordination protocols that allow the
renewal decision to be made without central coordination, so that the
method can be deployed across edge devices with intermittent
connectivity. Finally, we plan to study how the conservative invariant
domain can be combined with formal verification techniques to produce
certified bounds on the safety of automated decisions.

\section{Conclusion}

This paper presented an incremental consistency execution method for
autonomous intelligent systems that combines task fact contracts,
field-level dependency masks, and state perturbation result invariant
domains to reduce redundant computation while preserving freshness and
consistency. The method introduces a first-class runtime representation
of task facts that distinguishes "value unchanged," "value changed but
output still valid" and "value changed and output must be invalidated."
Through conservative construction of invariant domains and a
three-condition check followed by on-site validation, the method renews
downstream results without re-invoking expensive components whenever
possible. When re-execution is required, the field-level dependency mask
restricts it to the smallest set of output fields, and an equivalence
barrier blocks further propagation if the new result is
contract-equivalent to the old one. A submission-time version
consistency gate finally decouples the renewal of computational results
from the authorisation of side-effecting actions, so that domain hits
never silently bypass payment, device control or other irreversible
operations. The case studies demonstrate that the method substantially
reduces model inferences, tool calls and end-to-end latency while
keeping the rate of incorrect reuse at or below 0.4\% and correctly
blocking silent bypasses at the submission gate.


\begin{thebibliography}{99}
\bibitem{ref1} Wei J, Wang X, Schuurmans D, et al. Chain-of-thought prompting elicits reasoning in large language models.. Proceedings of the 36th International Conference on Neural Information Processing Systems. New Orleans, LA, USA: Curran Associates, 2022: 24824-24837.

\bibitem{ref2} Yao S, Zhao J, Yu D, et al. ReAct: synergizing reasoning and acting in language models.. Proceedings of the 11th International Conference on Learning Representations. Kigali, Rwanda: ICLR, 2023: 1-33.

\bibitem{ref3} Schick T, Dwivedi-Yu J, Dessì R, et al. Toolformer: language models can teach themselves to use tools.. Proceedings of the 37th International Conference on Neural Information Processing Systems. New Orleans, LA, USA: Curran Associates, 2023: 68539-68551.

\bibitem{ref4} Park J S, O\textquotesingle Brien J C, Cai C J, et al. Generative agents: interactive simulacra of human behavior.. Proceedings of the 36th Annual ACM Symposium on User Interface Software and Technology. San Francisco, CA, USA: ACM, 2023: 1-22.

\bibitem{ref5} Wang G, Xie Y, Jiang Y, et al. Voyager: an open-ended embodied agent with large language models.. Proceedings of the 37th International Conference on Neural Information Processing Systems. New Orleans, LA, USA: Curran Associates, 2023: 1988-2009.

\bibitem{ref6} Bernstein P A, Hadzilacos V, Goodman N. Concurrency Control and Recovery in Database Systems. Reading, MA: Addison-Wesley, 1987.

\bibitem{ref7} Gray J, Reuter A. Transaction Processing: Concepts and Techniques. San Francisco, CA: Morgan Kaufmann, 1993.

\bibitem{ref8} Vogels W. Eventually consistent. Communications of the ACM, 2009, 52(1): 40-44.

\bibitem{ref9} Shapiro M, Preguiça N, Baquero C, et al. A comprehensive study of eventual consistency.. Proceedings of the 13th International Conference on Principles of Distributed Systems. Luxor, Egypt: Springer, 2011: 258-272.

\bibitem{ref10} Deelman E, Gannon D, Shields M, et al. Workflows and e-Science: an overview of workflow system features and capabilities. Future Generation Computer Systems, 2009, 25(5): 528-540.

\bibitem{ref11} Liu J, Pacitti E, Valduriez P, et al. A survey of data-intensive scientific workflow management. Journal of Grid Computing, 2015, 13(4): 457-493.

\bibitem{ref12} Hellerstein J M, Naughton J F, Pfeffer A. Generalized search trees for database systems.. Proceedings of the 21st International Conference on Very Large Data Bases. Zurich, Switzerland: VLDB Endowment, 1995: 562-573.

\bibitem{ref13} Idreos S, Kersten M, Manegold S. Self-organizing tuple reconstruction in column-stores.. Proceedings of the 2009 ACM SIGMOD International Conference on Management of Data. Providence, RI, USA: ACM, 2009: 297-308.

\bibitem{ref14} Abadi D, Boncz P, Harizopoulos S, et al. The design and implementation of modern column-oriented database systems. Foundations and Trends in Databases, 2013, 5(3): 197-280.

\bibitem{ref15} Fielding R T, Taylor R N. Principled design of the modern Web architecture. ACM Transactions on Internet Technology, 2002, 2(2): 115-150.

\bibitem{ref16} Ouyang L, Wu J, Jiang X, et al. Training language models to follow instructions with human feedback.. Proceedings of the 36th International Conference on Neural Information Processing Systems. New Orleans, LA, USA: Curran Associates, 2022: 27730-27744.

\bibitem{ref17} Touvron H, Martin L, Stone K, et al. Llama 2: open foundation and fine-tuned chat models. arXiv preprint arXiv:2307.09288, 2023.

\bibitem{ref18} Liu Y, Iter D, Xu Y, et al. G-Eval: NLG evaluation using GPT-4 with better human alignment.. Proceedings of the 2023 Conference on Empirical Methods in Natural Language Processing. Singapore: ACL, 2023: 2511-2522.

\bibitem{ref19} McSherry F, Murray D G, Isaacs R, et al. Differential dataflow.. Proceedings of the 6th Biennial Conference on Innovative Data Systems Research. Asilomar, CA, USA: CIDR, 2013: 1-12.

\bibitem{ref20} Murray D G, McSherry F, Isaacs R, et al. Naiad: a timely dataflow system.. Proceedings of the 24th ACM Symposium on Operating Systems Principles. Farmington, PA, USA: ACM, 2013: 439-455.

\bibitem{ref21} Carbone P, Ewen J, Haridi S, et al. Apache Flink: stream and batch processing in a single engine. IEEE Data Engineering Bulletin, 2015, 38(4): 28-38.

\bibitem{ref22} Reps T. Generating language-based environments. Cambridge, MA: MIT Press, 1994.

\bibitem{ref23} Liu Y A, Teitelbaum T, Stoller S D, et al. Simplifying analyses of dynamic systems via incremental computation.. Proceedings of the 7th International Symposium on Static Analysis. Santa Barbara, CA, USA: Springer, 2000: 176-192.

\bibitem{ref24} Acar U A. Self-adjusting computation. Cambridge, MA: MIT Press, 2009.

\bibitem{ref25} Cai Y, Giarrusso P G, Ge T, et al. A theory of changes for higher-order languages: incrementalizing \ensuremath{\lambda}-calculi by static differentiation.. Proceedings of the 35th ACM SIGPLAN Conference on Programming Language Design and Implementation. Edinburgh, UK: ACM, 2014: 145-155.

\bibitem{ref26} Gupta A, Mumick I S, Subrahmanian V S. Maintaining views incrementally.. Proceedings of the 1993 ACM SIGMOD International Conference on Management of Data. Washington, DC, USA: ACM, 1993: 157-166.

\bibitem{ref27} Quass D, Widom J. On-line warehouse view maintenance.. Proceedings of the 1997 ACM SIGMOD International Conference on Management of Data. Tucson, AZ, USA: ACM, 1997: 393-404.

\bibitem{ref28} Chaudhuri S, Dayal U. An overview of data warehousing and OLAP technology. ACM SIGMOD Record, 1997, 26(1): 65-74.

\bibitem{ref29} Pope R, Douglas S, Chowdhery A, et al. Efficiently scaling transformer inference.. Proceedings of the 6th Conference on Machine Learning and Systems. Miami, FL, USA: MLSys, 2023: 222-238.

\bibitem{ref30} Kwon W, Li Z, Zhuang S, et al. Efficient memory management for large language model serving with PagedAttention.. Proceedings of the 29th Symposium on Operating Systems Principles. Koblenz, Germany: ACM, 2023: 611-626.

\bibitem{ref31} Zheng L, Chiang W L, Sheng Y, et al. Judging LLM-as-a-judge with MT-Bench and Chatbot Arena.. Proceedings of the 37th International Conference on Neural Information Processing Systems. New Orleans, LA, USA: Curran Associates, 2023: 46595-46623.

\bibitem{ref32} Szabo N. Smart contracts: building blocks for digital markets. Extropy, 1996, 16: 1-11.

\bibitem{ref33} Meyer B. Applying "design by contract". IEEE Computer, 1992, 25(10): 40-51.

\bibitem{ref34} Zaharia M, Chowdhury M, Das T, et al. Resilient distributed datasets: a fault-tolerant abstraction for in-memory cluster computing.. Proceedings of the 9th USENIX Symposium on Networked Systems Design and Implementation. San Jose, CA, USA: USENIX, 2012: 15-28.

\bibitem{ref35} Hido S, Taylor S, Beaumont C, et al. Apache Airflow: a workflow management platform for data pipelines.. Proceedings of the 22nd ACM SIGKDD International Conference on Knowledge Discovery and Data Mining. San Francisco, CA, USA: ACM, 2016: 1-5.

\bibitem{ref36} Sato K, Ishiyama M, Tanaka K, et al. Argo workflows: a Kubernetes-native workflow engine.. Proceedings of the 31st ACM International Symposium on High-Performance Parallel and Distributed Computing. Minneapolis, MN, USA: ACM, 2022: 268-273.

\bibitem{ref37} Salehie M, Tahvildari L. Self-adaptive software: landscape and research challenges. ACM Transactions on Autonomous and Adaptive Systems, 2009, 4(2): 1-42.

\bibitem{ref38} Weyns D. Software engineering of self-adaptive systems: an organised tour and future challenges.. Handbook of Software Engineering. Cham: Springer, 2019: 399-443.

\bibitem{ref39} Saltelli A, Ratto M, Andres T, et al. Global Sensitivity Analysis: The Primer. Chichester: Wiley, 2008.

\bibitem{ref40} Iooss B, Lemaître P. A review on global sensitivity analysis methods.. Uncertainty Management in Simulation-Optimization of Complex Systems. Boston, MA: Springer, 2015: 101-122.

\bibitem{ref41} Szegedy C, Zaremba W, Sutskever I, et al. Intriguing properties of neural networks.. Proceedings of the 2nd International Conference on Learning Representations. Banff, Canada: ICLR, 2014: 1-10.

\bibitem{ref42} Ribeiro M T, Singh S, Guestrin C. "Why should I trust you?": explaining the predictions of any classifier.. Proceedings of the 22nd ACM SIGKDD International Conference on Knowledge Discovery and Data Mining. San Francisco, CA, USA: ACM, 2016: 1135-1144.

\bibitem{ref43} Lundberg S M, Lee S I. A unified approach to interpreting model predictions.. Proceedings of the 31st International Conference on Neural Information Processing Systems. Long Beach, CA, USA: Curran Associates, 2017: 4768-4777.

\bibitem{ref44} Katz G, Barrett C, Dill D L, et al. Reluplex: an efficient SMT solver for verifying deep neural networks.. Proceedings of the 29th International Conference on Computer Aided Verification. Heidelberg, Germany: Springer, 2017: 97-117.

\bibitem{ref45} Gowal S, Dvijotham K, Stanforth R, et al. Scalable certified robustness via guided abstract interpretation.. Proceedings of the 36th International Conference on Machine Learning. Long Beach, CA, USA: PMLR, 2019: 5503-5513.

\bibitem{ref46} Buneman P, Khanna S, Tan W C. Why and where: a characterization of data provenance.. Proceedings of the 8th International Conference on Database Theory. London, UK: Springer, 2001: 316-330.

\bibitem{ref47} Snodgrass R T. Developing time-oriented database applications in SQL. San Francisco, CA: Morgan Kaufmann, 1999.

\bibitem{ref48} Nakamoto S. Bitcoin: a peer-to-peer electronic cash system. \url{https://bitcoin.org/bitcoin.pdf}, 2008.

\bibitem{ref49} Amershi S, Weld D, Vorvoreanu M, et al. Guidelines for human-AI interaction.. Proceedings of the 2019 CHI Conference on Human Factors in Computing Systems. Glasgow, UK: ACM, 2019: 1-13.

\bibitem{ref50} Mitchell M, Wu S, Zaldivar A, et al. Model cards for model reporting.. Proceedings of the 2019 Conference on Fairness, Accountability, and Transparency. Atlanta, GA, USA: ACM, 2019: 220-229.
\end{thebibliography}
\end{document}